# Center-Focusing Multi-task CNN with Injected Features for Classification of Glioma Nuclear Images

Veda Murthy
Lexington High School
murthyveda0@gmail.com

Le Hou
Stony Brook University
le.hou@stonybrook.edu

Dimitris Samaras
Stony Brook University,
Ecole Centrale Paris
samaras@cs.stonybrook.edu

Tahsin M. Kurc
Stony Brook University,
Oak Ridge National Laboratory
tahsin.kurc@stonybrook.edu

Joel H. Saltz
Stony Brook University,
Stony Brook University Hospital
joel.saltz@stonybrookmedicine.edu

## Abstract

*Classifying the various shapes and attributes of a glioma cell nucleus is crucial for diagnosis and understanding of the disease. We investigate the automated classification of the nuclear shapes and visual attributes of glioma cells, using Convolutional Neural Networks (CNNs) on pathology images of automatically segmented nuclei. We propose three methods that improve the performance of a previously-developed semi-supervised CNN. First, we propose a method that allows the CNN to focus on the most important part of an image- the image's center containing the nucleus. Second, we inject (concatenate) pre-extracted VGG features into an intermediate layer of our Semi-Supervised CNN so that during training, the CNN can learn a set of additional features. Third, we separate the losses of the two groups of target classes (nuclear shapes and attributes) into a single-label loss and a multi-label loss in order to incorporate prior knowledge of inter-label exclusiveness. On a dataset of 2078 images, the combination of the proposed methods reduces the error rate of attribute and shape classification by 21.54% and 15.07% respectively compared to the existing state-of- the-art method on the same dataset.*

## 1. Introduction

Brain tumors are the most common cause of cancer-related death among people ages 0-19 and the second leading cause of cancer-related death in children. While there are over 100 distinct types of brain tumors, Gliomas constitute about 27% of all brain tumors and 80% of all malignant tumors [3]. There are several subtypes of glioma, each of which requires a different form of treatment. The most common way to diagnose and differentiate which subtype of glioma a patient has is through examination of a simple hematoxylin and eosin stained brain biopsy slide. A pathologist analyzes the cells of the slide, looking for specific attributes of each cell and nucleus and the shape of the nucleus. The combination of the two will determine the possible type of glioma a patient may have. The attributes and shapes a pathologist looks for are primarily based off guidelines given by the World Health Organization Classification of Tumors of the Central Nervous System [14, 15]. While manual analysis of histological biopsy images has been commonplace for over a century, automation has the inherent advantage of both reproducibility and the lack of individualized qualitative judgement by the examining pathologist [11]. Modern Slide scanners now have the ability to produce a digitized, gigapixel Whole Slide Image (WSI) of a given biopsy. Because a single whole slide image typically contains about one hundred million nuclei, pathologists cannot examine all nuclei carefully for diagnosis. Automated classification of the attributes of a given cell and nucleus will allow pathologists to quickly gain access to specific information needed to determine the glioma subtype and develop targeted treatment plans.

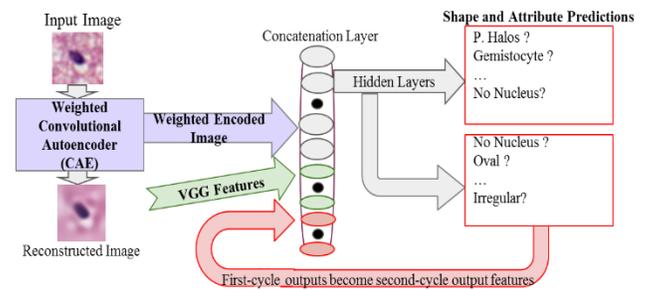

Figure 1: Summary of the Approach. Our proposed methods are in boldface, and include the weighting of the central pixels of the image in the CAE, the addition of pre-extracted VGG features to a semi-supervised CNN, and the separate single and multi-label loss functions.

Automated attribute classification for cells in histology images has been studied before. However, past research classified cells into fewer categories. For example, one approach [22] recognizes only four attributes in nuclei; another [25] only recognizes healthy vs pathological nuclei. A recent work [8] focuses on nine nuclear visual attributes. However, in practice, at least fifteen nuclear shapes and visual attributes are needed in order to classify subtypes of glioma. Limited attribute recognition inherently limits the information a pathologist can receive through automated classification, thereby reducing the ability to accurately determine the glioma subtype. We use an expanded number of labels in our classifier- nine non-mutually exclusive attributes and six mutually exclusive shapes. These attributes were selected by a pathologist and are very important for diagnosis and treatment purposes.

We use Convolutional Neural Networks (CNNs) to classify nuclear attributes. CNNs have proven themselves to be state-of-the-art algorithms in both common image classification [7, 9] and medical image analysis [4]. They have achieved error rates below that of human classification in the ImageNet classification challenge [7] and have posted state-of-the-art results in the MICCAI mitosis detection challenge [4], as well as whole slide glioma classification [28]. The promising results of CNNs on other datasets was the primary reason for adopting them for our research.

Fully supervised CNNs require large, labeled datasets. As a result, many of the datasets where CNNs have performed the best have several thousand if not millions of labeled images. However, datasets with tens of thousands of labeled images are often out of reach for medical applications, where image annotation often requires pathologists with years of professional training. Because of this, effective usage of the available labeled data is a necessity in medical image applications. This paper explores innovative methods to train a CNN with a limited labeled dataset.

Recently, advances have been made in terms of using CNNs to classify glioma nuclei. In [8], the authors proposed two promising methods for glioma attribute classification. In one method, the activations from the last fully-connected layer of a VGG-16 network [21] are used to train a support vector machine. The VGG-16 CNN was trained on ImageNet images, and the features were extracted from glioma nuclear images [6, 18]. The other method is a semi-supervised CNN, where the convolutional and pooling layers of the CNN are pretrained using a convolutional autoencoder (CAE) [17] on unlabeled nuclear images. Later hidden layers are trained in an exclusively supervised manner. To learn inter-attribute correlations between the features, the CNN is trained in two cycles. In the second cycle, output probabilities for each of the labels from the previous cycle are used to help train the CNN, allowing the CNN to see correlations between attributes [20]. We retain all of these advances, and improve upon them using novel methods of our own.

**Our first contribution** is to weight the autoencoder's loss function to favor reconstruction of central image pixels, because each image is exclusively labeled in terms of shapes and attributes for the centermost nuclei of the image. In pathology images, nuclei can be close to, or overlapping on each other. Therefore, it is inevitable that the autoencoder's input image may contain multiple nuclei. The additional, non-central nuclei will pose a source of noise to the classification stage. Thus, pretraining the CNN with an autoencoder to focus on the image center will reduce the chance of severe overfitting.

It has been shown that combining CNNs with features extracted by other models can improve classification results [10, 17, 23]. Therefore, we seek to inject (concatenate) VGG-16 activations into the semi-supervised CNN to provide additional information to the CNN. As **our second contribution**, the VGG-16 features are added into the CNN during training. Therefore, the CNN is aware of the added features and can learn a set of additional features.

**Our third contribution** is to expand the CNN of [8], to include vital shape information in terms of outputs. Nuclear shapes are critical for discriminating lower-grade Astrocytoma which tend to have elongated nuclei, and Oligodendroglioma which tend to have round nuclei [2]. Note that a nucleus can have multiple visual attributes but can only belong to one shape category (elongated, rounded, etc.). We model this inter-label exclusivity through single-label and multi-label tasks for shapes and attributes respectively. The loss functions for these two tasks are subsequently combined into a single loss function, used for backpropagation on a single CNN. This can be viewed as a multi-task CNN [5, 7]. In our CNN architecture, the convolutional, pooling and first hidden layer of shapes and attributes are shared, while shapes and attributes have separate hidden and output layers [1, 27].

We conducted experiments on the dataset of 2078 nuclear images from [8]. Using the same training and testing set separation, our combined method achieved a promising 0.8570 Area under ROC curve (AuROC), reducing the error of the existing state-of-the-art method [8] by 19%.

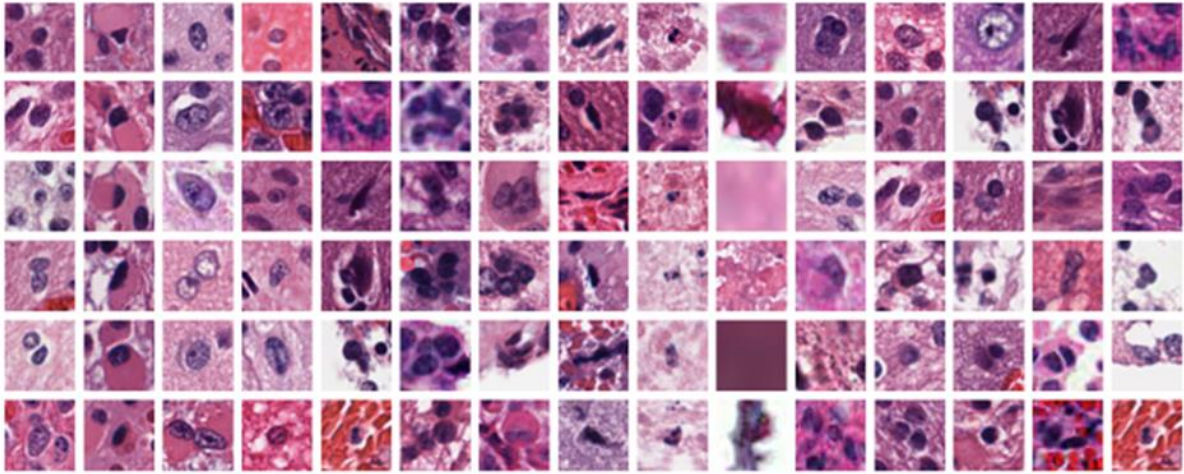

Figure 2: Examples of attributes and shapes. Starting from the leftmost column: Perinuclear Halos, Gemistocyte, Nucleoli, Grooved, Hyperchromasia, Overlapping Nuclei, Multinucleation, Mitosis, Apoptosis, No Nucleus, Oval Nucleus, Close to Round Nucleus, Round Nucleus, Elongated Nucleus and Irregular Shaped Nucleus.

## 2. Nuclear attribute and shape recognition

The visual attributes and shape of nuclei are fundamental in cancer subtype diagnosis. We apply our method on a dataset that is expanded based on an existing dataset [8] with additional nuclear shape classes. There are 2078 RGB images of nuclei extracted from Hematoxylin-Eosin stained tissue images. Each of these 2078 images is already labeled in a binary manner for nine important visual attributes. Because the nuclei were automatically segmented, several images are mis-segmentations that contain no nucleus- thus the dataset has the no nucleus category which acts as a mutually exclusive attribute. The magnification of nuclear images is 20X (0.5 microns per pixel) and the resolution is 50 by 50 pixels.

We then expand this dataset to include five shape classes. These mutually-exclusive shape classes were created by a pathologist and a trained graduate student. In the case that there were multiple nuclei in an image, the labels correspond to the nucleus closest to the image center. In addition, our supervised CNN was pretrained in an unsupervised fashion as a Convolutional Autoencoder (CAE) using 150,000 unlabeled images of nuclei.

## 3. Our Methods

In this section we introduce our three proposed methods. Using these methods, we significantly improved the mean Area under the ROC curve (AuROC) by the previous semi-supervised CNN [8].

| Attribute / Shape | #. Positives | #. Negatives |
|---|---|---|
| Perinuclear Halos | 78 | 2000 |
| Gemistocyte | 51 | 2027 |
| Nucleoli | 77 | 2001 |
| Grooved | 14 | 2064 |
| Hyperchromasia | 505 | 1573 |
| Overlapping Nuclei | 105 | 1973 |
| Multinucleation | 43 | 2035 |
| Mitosis | 53 | 2025 |
| Apoptosis | 20 | 2058 |
| No Nucleus | 545 | 1533 |
| Oval | 325 | 1753 |
| Close to Round | 104 | 1974 |
| Round | 296 | 1782 |
| Elongated | 333 | 1745 |
| Irregular Shape | 475 | 1603 |

Table 1: This table shows the number of positive and negative examples in the 2078 image dataset used. Note that many images have more than one attribute, and that positive examples in some categories, such as in grooved, are limited.

### 3.1. CNN initialized to focus on image center

In semi-supervised learning, a Convolutional Autoencoder (CAE) is used to initialize the supervised CNN. The loss function of a conventional CAE is the Mean Squared Error (MSE) between input image pixels and reconstructed image pixels. The reconstruction error of pixels in any location of the input images influence the loss in the same way. However, in many applications, part of the image, usually the center of the image, is more important than other parts. We propose to weight the loss function to focus on the reconstruction of the image centers. In particular, the reconstruction errors of the center $c \times c$ pixels are weighted $w$ times. The resulting Weighted Mean Squared Error (WMSE) is expressed as:

$$L_{CAE}(X, R) = \sum_{i,j=1}^{d} W_{i,j}(R_{i,j} - X_{i,j})^2, \quad (1)$$

where $X$ and $R$ are input and reconstructed images (matrices) of size $d$ respectively, and $W$ is the weight matrix. In this paper we use the following weight matrix:

$$\begin{aligned} W_{i,j} &= w & \text{If } i, j \in (\tfrac{d}{2} - \tfrac{c}{2}, \tfrac{d}{2} + \tfrac{c}{2}) \\ W_{i,j} &= 1 & \text{Otherwise.} \end{aligned} \quad (2)$$

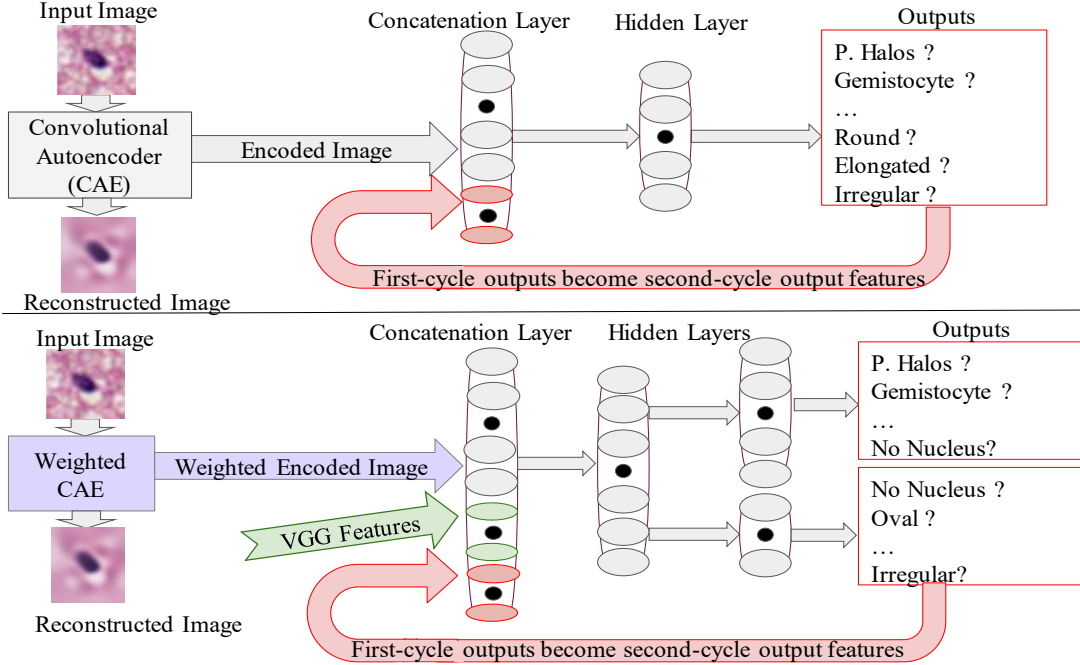

Figure 3: (Top) The Default-CNN used in [8]. (Bottom) The WFM-CNN, which uses the weighted CAE loss function, merged single and multi-label losses and pre-extracted features.

In our experiments, we set $w = 5$ and $c = 20$. After training the CAE with WMSE, we initialize our CNN with the CAE. We name the CNN initialized with this CAE weighted CNN (W-CNN).

### 3.2. Training a CNN with additional features

We propose to extract features from a pretrained CNN and inject (concatenate) them into our semi-supervised CNN. We call this method Feature Concatenation-CNN (F-CNN). Because our semi-supervised CNN is trained with these additional features, it can learn a set of features in addition to those of the pretrained CNN.

In particular, VGG features from the first fully-connected layer of a pretrained VGG 16-layer network [20] are concatenated to the concatenation layer (See Fig. 3). In order to reduce the dimensionality of the concatenation layer, we add a fully connected layer of 1000 units between the concatenation layer and the second-to-last fully connected layer of 100 units.

### 3.3. Separating single and multi-label tasks

We expanded the previous dataset [8] to include nuclear shape classes. In contrast to other visual attributes, shapes are mutually exclusive (an elongated nucleus cannot be round at the same time). Modeling the problem of classifying all visual attributes with shape classes as a multi-label problem does not utilize this prior knowledge. We propose to separate the output layer and also the layer prior to the output layer, to learn nuclear shapes and attributes separately. The attribute prediction branch uses the sigmoid output activation function, whereas the shape prediction branch uses the softmax output activation function.

The losses of the two branches are computed as a multi-label log-likelihood $L_{ml}$ and single-label log-likelihood $L_{sl}$

separately. We finally merge those two losses by a weighted sum:

$$L_{CNN} = m \times L_{ml} + (1-m) \times L_{sl}, \quad (3)$$

where $m$ is the combining weight. In this paper we use $m = 0.6$. We name the CNN trained with the loss represented by Eq. 3 multi-loss CNN (M-CNN). Other methods in this paper including the baseline models the nuclear attribute and shape classification simply as one multi-label classification problem.

## 4. Experiments

We conducted experiments using the proposed methods and showed significant improvements compared to the previous state-of-the art method [8]. We achieved an average AuROC of 0.8570 across all classes, which reduced the error by 19% on the dataset we tested.

### 4.1. Implementation details

This section discusses implementation details. The architecture of our CNN (shown below) is the same as the previous approach [8].

| Layers | Output size | Note |
|---|---|---|
| Input image | 3 x 32 x 32 | RGB images |
| Dropout | 3 x 32 x 32 | p = 0.05 |
| Conv | 80 x 30 x 30 | Filter size: 3 |
| Conv | 80 x 28 x 28 | Filter size: 3 |
| Conv | 120 x 26 x 26 | Filter size: 3 |
| Maxpool | 120 x 13 x 13 | Pool size: 2, stride: 2 |
| Conv | 100 x 11 x 11 | Filter size: 3 |
| Conv | 140 x 9 x 9 | Filter size: 3 |
| Conv | 140 x 7 x 7 | Filter size: 3 |
| Maxpool | 140 x 3 x 3 | Pool size: 2, stride: 2 |
| Fully-Con. | 400 | |
| Fully-Con. | 100 | |
| Concatenation | 119 | Concate 1st cycle predictions |
| Fully-Con. | 100 | |
| Sigmoid | 19 | |

Table 2: The architecture of the Default-CNN. Note that the ReLU activations have been removed for simplicity of the table.

Our CNN was implemented using Theano [23] and Lasagne [27]. VGG features were extracted through a pretrained VGG-CNN network in Matconvnet [25]. We trained our CNN on a single Nvidia GeForce GTX Titan X GPU. We used stochastic gradient descent with momentum as the optimization method for all CAEs and CNNs. For each method, we set the momentum at 0.975 and selected the best learning rate individually. Training a CAE took approximately 12 hours. Training and testing a CNN initialized by CAE for five random-split validations took around three hours. The tested methods were:

1. **Default-CNN** [8]. We used the exact code of the authors of [8]. This approach models the nuclear attribute and shape classification simply as a single multi-label classification problem. The loss used is binary cross-entropy (log-likelihood on each attribute and shape class). The authors of [8] initialized the default-CNN using a conventional CAE. The learning rate of the CNN was 0.0005.

2. **W-CNN**: This the weighted CNN that focuses on image centers (Sec. 3.1). We initialized the CNN with a CAE that used a Weighted Mean Squared Error (WMSE) as its loss function. This is the only difference between W-CNN and default-CNN. The learning rate we used was 0.0005.

3. **WF-CNN**: The combination of the W-CNN (Sec. 3.1) and F-CNN (Sec. 3.2) methods. F-CNN uses pre-extracted VGG16 features as additional injected features. The learning rate we used was 0.0001. different from the Default-CNN because of the injected features.

4. **WFM-CNN**: The combination of the W-CNN, F-CNN and M-CNN (Sec 3.3) methods. The M-CNN separates the shape and attribute classification tasks. The learning rate was 0.0001 and reduced by a factor of 10 for every 50 epochs. These changes were made because of the added shape classes. We also emphasize that "no nucleus" is considered both a shape and attribute for the purposes of the WFM-CNN- thus it must be both a sigmoid output and a softmax output.

5. **Combined WFM-CNN / WF-CNN**: We use the WFM- CNN to predict nuclear shapes, and the WF-CNN to predict nuclear visual attributes. The WFM-CNN is able to model the mutual exclusivity of shape classes. Therefore, we expect it to achieve the best result recognizing nuclear shapes. However, the WF-CNN can achieve the better results on classifying attributes, because features in the second-to-last- layer are shared for both shape classification and attribute classification (Figure 2). Therefore, features used for shape classification are also included for the attribute classification branch. This combination of two CNNs is able to accomodate treating shapes as a special type of label, while allowing attributes to remain unaffected by CNN changes designed for shape labels.

| | Default-CNN [8] | Combined WF-CNN / WFM-CNN |
|---|---|---|
| Error Rate on Attr. | 0.1513 | 0.1187 (**21.54%** ↓) |
| Error Rate on Shapes | 0.2269 | 0.1927 (**15.07%** ↓) |

Table 3: The error rates of the Default-CNN [8] and our combined WF-CNN / WFM-CNN. Our method shows substantial decreases in error rate.

| Attributes and Shapes | Methods of [8] | Our Methods | | | |
|---|---|---|---|---|---|
| | | W-CNN | WF-CNN | WFM-CNN | Combo of WF-CNN and WFM-CNN |
| Perinuclear Halos | 0.8316 | 0.8312 | **0.9171** | 0.9041 | **0.9171** |
| Gemistocyte | 0.839 | 0.842 | **0.9363** | 0.8809 | **0.9363** |
| Nucleoli | 0.8793 | 0.9002 | **0.9215** | 0.901 | **0.9215** |
| Grooved | 0.6566 | **0.7731** | 0.7145 | 0.6223 | 0.7145 |
| Hyperchromasia | **0.9514** | 0.9487 | 0.9244 | 0.9142 | 0.9244 |
| Overlapping Nuclei | 0.8755 | 0.875 | **0.9118** | 0.8896 | **0.9118** |
| Multinucleation | 0.7239 | **0.7805** | 0.7402 | 0.7083 | 0.7402 |
| Mitosis | **0.911** | 0.8946 | 0.8822 | 0.8293 | 0.8822 |
| Apoptosis | 0.8388 | 0.8548 | **0.8837** | 0.8721 | **0.8837** |
| No Nucleus | 0.9795 | 0.9781 | **0.9816** | 0.9709 | **0.9816** |
| Oval | 0.6569 | 0.6596 | 0.7022 | **0.7254** | **0.7254** |
| Close to Round | 0.7565 | 0.7397 | 0.749 | **0.7681** | **0.7681** |
| Round | 0.8434 | 0.834 | 0.8633 | **0.8726** | **0.8726** |
| Elongated | 0.8634 | 0.8644 | 0.8796 | **0.8855** | **0.8855** |
| Irregular | 0.7455 | 0.7601 | 0.7647 | **0.7902** | **0.7902** |
| Mean AuROC | 0.8235 | 0.8357 | 0.8515 | 0.8316 | **0.857** |
| Mean AuROC for Shapes Alone | 0.7731 | 0.7716 | 0.7918 | **0.8083** | **0.8083** |

Table 4: This chart shows the breakdown of ROC values by attribute and shape, and by method. Overall, the combined WF-CNN / WMF-CNN employing all three proposed improvements (weighted CNN, multi-loss functions and added VGG features) was the most accurate for all shapes and the most accurate for the majority of attributes.

## 5. Results

We trained all CNNs on approximately 1600 labeled images, and reserved 400 images for testing. We repeated this process five times and averaged the results. Splits between training and testing datasets were random though reproducible- we used a fixed seed for random number generation in all experiments. Thus, all CNNs in each fold had the same training/testing split. We use the Area under ROC curve (AuROC) as the evaluation metric. We show the ROC curves in Fig. 4.

The results are shown in Table 4. Our combined WF-CNN/ WFM-CNN outperforms the Default-CNN for the majority of attributes and all shapes in terms of AuROC. Our combined CNN also has a lower error rate, decreasing the error rate by 21.54% on attributes and 15.07% on shapes, shown in Table 3. Note that the results of the Default-CNN are slightly different in the original paper [8], because of unknown training and testing splits in the original paper. In addition, the results of the original paper [8], only contain results for nuclear attributes, and not nuclear shapes.

## 6. Conclusions

We focused on recognizing visual attributes and shapes of nuclei in glioma histopathology images. This kind of histopathology datasets are relatively small because gathering ground truth labels require pathologists. Therefore, we used a semi-supervised CNN for this task. We proposed three methods: First, we concatenated pre-extracted VGG features into an intermediate layer of our CNN. Therefore, our CNN can learn a set of additional features during the training phase. Second, we created a CNN with two separate output layers. One output layer predicts mutually exclusive classes (shapes of nuclei) and the other output layer predicts non-mutually exclusive classes (attributes of nuclei). Third, we weighted the loss function of the CAE that initialized our CNN, to force the encoded features to better represent nuclei that are in the center of images. All methods were proven to be effective in terms of increasing CNN performance compared to the baseline CNN that had so far achieved state-of-the-art results on the dataset.

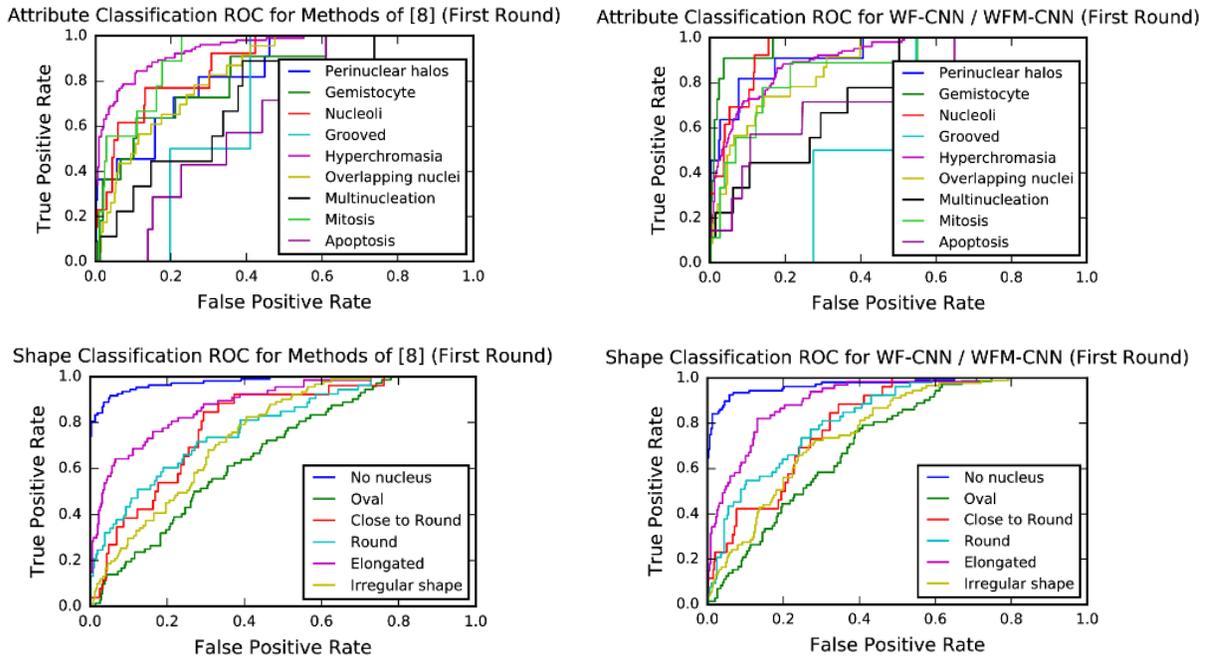

Figure 4. ROC Curves of the first training/validation round for nuclear visual attribute prediction (Top) and nuclear shape prediction (Bottom). Our WF-CNN/WFM-CNN outperformed the default CNN [8] significantly in most classes.

## Acknowledgements

This work was supported in part by 1U24CA180924-01A1 from the NCI, and R01LM011119-01 and R01LM009239 from the NLM as well as FRA DTFR5315C00011, the Subsample project from DIGITEO Institute, France, and a gift from Adobe Corp.